\documentclass[10pt, twocolumn, letterpaper]{article}

\usepackage{cvpr}              % To produce the CAMERA-READY version
\usepackage{graphicx}
\usepackage{amsmath}
\usepackage{amssymb}
\usepackage{booktabs}

\usepackage{times}
\usepackage{epsfig}
\usepackage{bm}
\usepackage{subcaption}
\usepackage{url}
\usepackage{gensymb}
\usepackage{ulem}
\usepackage[symbol]{footmisc}
\usepackage{makecell}
\usepackage{xcolor}
\usepackage{nopageno}
\usepackage{float}
\usepackage{tabularx}
\usepackage[pagebackref,breaklinks,colorlinks]{hyperref}

\usepackage[capitalize]{cleveref}
\crefname{section}{Sec.}{Secs.}
\Crefname{section}{Section}{Sections}
\Crefname{table}{Table}{Tables}
\crefname{table}{Tab.}{Tabs.}

\ifdefined\siggraph
\usepackage{times}
\fi

\usepackage{color}
\usepackage{ifthen}
\usepackage{float}
\usepackage{alltt}
\usepackage{newlfont} % for Box
\usepackage{array}

\usepackage{wrapfig}
\usepackage{booktabs}
\usepackage{multirow}

\newcommand{\PreserveBackslash}[1]{\let\temp=\\#1\let\\=\temp}
\newcommand{\Caption}[2]{\caption[#1]{{\footnotesize #1} {\footnotesize #2}}}

\definecolor{DeltaColor}{rgb}{0.039,0.73,0.71}
\definecolor{SetaColor}{rgb}{0.867, 0.0235, 0.376}
\definecolor{SigmaColor}{rgb}{0.98,0.45,0.0}
\definecolor{RedColor}{rgb}{0.8,0,0}
\definecolor{AlphaColor}{rgb}{0,0,0.8}
\definecolor{BetaColor}{rgb}{0.8,0,0.8}
\definecolor{GammaColor}{rgb}{0.5,0,0.7}
\definecolor{EpsilonColor}{rgb}{0.353,0.725,0.906}
\definecolor{TauColor}{rgb}{0.423,0.235,0.192}

\newcommand{\nothing}[1]{}

\definecolor{AudioColor}{rgb}{0.56,0.34,0.62}

\definecolor{DeadlineColor}{rgb}{0.9,0.4,0} % energetic color

\definecolor{figred}{rgb}{1,0,0}
\definecolor{figgreen}{rgb}{0,0.6,0}
\definecolor{figblue}{rgb}{0,0,1}
\definecolor{figpink}{rgb}{1,0.63,0.63}

\newcounter{pccount}
\floatstyle{plain}
\newcommand{\filename}[1]{\url{#1}}
\newcommand{\foldername}[1]{\url{#1}}

\newcommand{\distribution}{\mathbb{P}}
\newcommand{\realx}{\bm{x}}

\newcommand{\coords}{\bm{c}}
\newcommand{\guidance}{\bm{g}}
\newcommand{\encodingFunc}{\gamma}

\def\cvprPaperID{7466} % *** Enter the CVPR Paper ID here
\def\confName{CVPR}
\def\confYear{2022}

\graphicspath{
	{figs/raster/}
	{figs/handdrawn/}
}

\begin{document}
%\title{Implicit Texture Field Network with Periodic Encoding}
\title{Exemplar-based Pattern Synthesis with Implicit Periodic Field Network}

% The \author macro works with any number of authors. There are two commands
% used to separate the names and addresses of multiple authors: \And and \AND.
%
% Using \And between authors leaves it to LaTeX to determine where to break the
% lines. Using \AND forces a line break at that point. So, if LaTeX puts 3 of 4p
% authors names on the first line, and the last on the second line, try using
% \AND instead of \And before the third author name.

	\author{Haiwei Chen$^{1,2}$, Jiayi Liu$^{1,2}$, Weikai Chen$^{3}$, Shichen Liu$^{1,2}$, Yajie Zhao$^{1,2}$ \\
		$^1$University of Southern California\\
		$^2$USC Institute for Creative Technologies\\
		$^3$Tencent Game AI Research Center\\
		{\tt\small \{haiweich,jliu8328\}@usc.edu \tt chenwk891@gmail.com \tt \{lshichen,zhao\}@ict.usc.edu  }
		% For a paper whose authors are all at the same institution,
		% omit the following lines up until the closing ``}''.
		% Additional authors and addresses can be added with ``\and'',
		% just like the second author.
		% To save space, use either the email address or home page, not both
% 		\and
% 		Second Author\\
% 		Institution2\\
% 		First line of institution2 address\\
% 		{\tt\small secondauthor@i2.org}
	}

\maketitle

%%%%%%%%% ABSTRACT
\begin{abstract}

Synthesis of ergodic, stationary visual patterns is widely applicable in texturing, shape modeling, and digital content creation. The wide applicability of this technique thus requires the pattern synthesis approaches to be scalable, diverse, and authentic. In this paper, we propose an exemplar-based visual pattern synthesis framework that aims to model the inner statistics of visual patterns and generate new, versatile patterns that meet the aforementioned requirements. To this end, we propose an implicit network based on generative adversarial network (GAN) and periodic encoding, thus calling our network the Implicit Periodic Field Network (IPFN). The design of IPFN ensures scalability: the implicit formulation directly maps the input coordinates to features, which enables synthesis of arbitrary size and is computationally efficient for 3D shape synthesis. Learning with a periodic encoding scheme encourages diversity: the network is constrained to model the inner statistics of the exemplar based on spatial latent codes in a periodic field. Coupled with continuously designed GAN training procedures, IPFN is shown to synthesize tileable patterns with smooth transitions and local variations. Last but not least, thanks to both the adversarial training technique and the encoded Fourier features, IPFN learns high-frequency functions that produce authentic, high-quality results. To validate our approach, we present novel experimental results on various applications in 2D texture synthesis and 3D shape synthesis. 

\end{abstract}

% \vspace{50mm}
% \input{figures/fig_teaser}

\section{Introduction}
\label{sec:intro}

The synthesis of visual patterns, may that be a wooden texture for painting, or a simulation of natural cave systems, is a technique that is applied ubiquitously in computer-aided design and digital content creation. Visual patterns can be understood as arts, shapes, or natural textures following certain geometric structures. In an application context, let us start by defining several characteristics that are desirable for an algorithm that generates visual patterns:

\begin{itemize}
    \item \textit{Authenticity}. Probably the most prioritized quality of synthesized visual patterns is its visual quality. When patterns are synthesized from an exemplar, the quality is determined by whether they faithfully recreate the source pattern.
    \item \textit{Diversity}. It would be undesirable for a synthesizer to only copy patterns from the source. Diversity is thus an equally important measurement that evaluates whether the synthesized patterns vary from the source and each other. We strive to achieve two different levels of diversity: the patterns should be diversified both \textit{within} a generated sample and \textit{across} samples.
    \item \textit{Scalability}. As patterns are usually demanded at different and potentially large scales for many practical applications, we want a scalable synthesizer to be able to efficiently generate patterns of arbitrary size. Scalability is particularly valuable when it comes to the synthesis of 3D models, as the extra dimension translates to a much larger amount of computations. 
\end{itemize}

A \textit{scalable} design choice leads us to formulate the synthesis problem as generating patterns from a continuous, real coordinate space. This is generally known as the implicit formulation, where a nonlinear function maps points defined in $\mathbb{R}^2$ or $\mathbb{R}^3$ to features that represent the synthesized subjects. In particular, the implicit function has been shown to be an efficient representation for synthesizing 3D volumes~\cite{henzler2020learning,park2019deepsdf,mildenhall2020nerf}. 

Patterns that scale well to an infinitely large space, in general, possess a stationary property - a shift-invariant structure that can be expanded by tiling or stacking blocks of elements. We therefore develop our method by pivoting on the fact that many types of natural and artistic patterns can be analyzed and recreated in a stationary framework. The goal of synthesizing an \textit{ authentic} and \textit{diverse} stationary pattern from an exemplar, however, requires careful modeling that is compatible with the underlying structure of the pattern.

\begin{figure}[!t]
 \begin{center}
  \centering
  \includegraphics[width=0.6\linewidth]{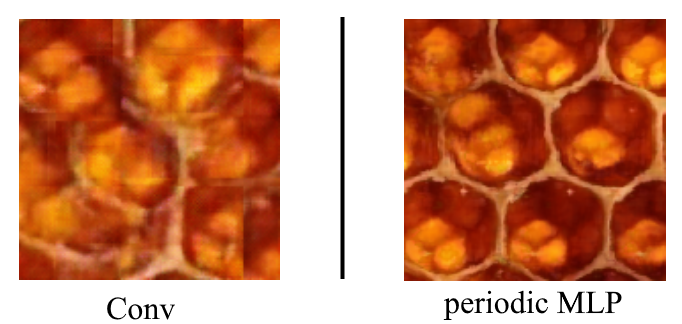}
 \end{center}
 \vspace{-6mm}
 \Caption{Comparison between synthesized honeycomb from a DCGAN convolution generator and the periodic MLP generator. Seams and intercepting patterns are visible in the former result due to difficulty for the convolution generator to capture the repeating structure.}{}
 \label{fig:comparison}
\end{figure}

Generative adversarial networks (GAN)~\cite{goodfellow2014generative} is one of the most promising techniques so far to model data distribution in an unsupervised manner and has been frequently adapted to convolutional models that synthesize visually authentic images~\cite{shaham2019singan,jetchev2016texture,bergmann2017learning,zhou2018non,xian2018texturegan,shocher2019ingan}. How would a GAN generator be leveraged to modeling stationary pattern? As all stationary patterns contain a repeating structure with local variations that ``vivifies'' its appearance, in an ideal situation, a stationary pattern can be modeled by a discrete random field, where each random variable is associated with a patch of the basic element. Thus a natural GAN formulation models image patches with a spatially defined latent field~\cite{jetchev2016texture,bergmann2017learning}. In a convolutional framework, however, problems arise when fake samples generated from a discrete noise image are discriminated from randomly sampled patches from a real image. The first problem is that the sampled patch does not necessarily agree with the scale of the repeating structure. The second problem is that the sampled patch can be arbitrarily shifted from the center of a stationary element. A typical deconvolutional network~\cite{dong2015image,zeiler2010deconvolutional} that upsamples from an evenly-spaced noise image may not sufficiently address the previously mentioned problems. To study their effects we designed a convolutional network following the DCGAN architecture~\cite{radford2015unsupervised} to synthesize a honeycomb pattern from a $2 \times 2$ noise map, which is trained against random patches sampled from a source image. The comparison between its result and that synthesized by our generator network that is trained with an identical discriminator is shown in Figure.~\ref{fig:comparison}. We found that the noise map does not capture well the honeycomb structure as seams and intercepting elements are visible from the synthesized image.

Though various techniques have been proposed in the past to address the aforementioned issues (e.g.~\cite{jetchev2016texture,bergmann2017learning}), in this paper, we consider a more natural way to model stationary patterns with an implicit periodic generator. The core of the formulation is to match the repeatable structure of a stationary pattern to the period of a learnable continuous periodic function. Instead of modeling the pattern with a discrete noise tensor, we define latent variables in a continuous space, where the extent of each latent factor is learned to match with the extent of the repeating elements. The benefits of this design align well with the desirable characteristics for visual pattern synthesis: 1) learned periodicity of the implicit field encourages latent factors to model the stationary variations observed from the exemplar pattern; 2) a continuous representation provides flexibility during training to learn a distribution from randomly shifted patches cropped from the exemplar; 3) a Fourier encoding scheme learns high-frequency details from the exemplar. This allows our model to synthesize visually authentic results, and 4) multilayer perceptron (MLP) that implicitly takes coordinates as input scales well to the generation of 3D shapes when compared to 3D convolution. Based on these design choices, we term our network the Implicit Periodic Field Network (IPFN).

We validate our proposed design by showing various applications in texture image synthesis and 3D volume synthesis in Section.~\ref{sec:result}. Specifically, besides synthesizing stationary patterns, we design a conditional formulation of our model to tackle the synthesis of directional patterns and to provide application in controllable shape generation. An ablation study is also conducted to verify the effectiveness of our design choices.

\section{Related Work}
\label{sec:related_work}

% Based on application: deep exemplar model, texture network, traditional shape modeling
% Based on method: implicit network, nerf-like
% Based on training: spatial GAN structure, GAN for image synthesis and 3D shape

\paragraph{Pattern Synthesis}

% ``Pattern" is a loosely defined term that is not restricted to the commonly studied texture pattern. In our work we extend from texture synthesis to the synthesis of ``patterned" 3D shapes, and thus in this section we cover synthesis algorithms in both the 2D and 3D realm. 

2D visual patterns are generally referred to as ``texture" due to their prevalent applications in computer-aided design.  Well-known early attempts to synthesize texture derive patterns from smoothly interpolated noise~\cite{perlin1985image,perlin2002improving,worley1996cellular} and create aesthetically pleasing materials that display a high level of randomness. \cite{henzler2020learning,portenier2020gramgan} are two recent works related to us that utilize randomness from a continuously-defined Perlin noise~\cite{perlin1985image} to synthesize exemplar-based textures in an implicit field. Their works demonstrate the advantage of smooth noise field and implicit formulation in efficiently and diversely generating a 3D texture field. However, just like their procedural precedence, these methods have been shown to be limited in synthesizing patterns that are more complicated in structures (e.g. bricks, rocky surface)~\cite{portenier2020gramgan}.       

Later works in traditional image synthesis are generally divided into pixel-based method (e.g.~\cite{efros1999texture,wei2003texture}), patch-based method (e.g.~\cite{kopf2007solid,efros2001image, kwatra2003graphcut}) and optimization-based method~(e.g.~\cite{kopf2007solid,kwatra2005texture}) and have shared with us important considerations for recreating new patterns from an exemplar. For instance, synthesis on a patch level encourages preservation of fine local details, and the efforts are focused on "quilting" the discontinuity between patches~\cite{efros2001image,kwatra2003graphcut} and encouraging global similarity~\cite{kopf2007solid}. Early statistical model~\cite{efros1999texture, portilla2000parametric} utilizes a random field representation that captures variations in stationary patterns and synthesizes a variety of new patterns under the Julesz' conjecture. Our work is inspired by key ideas from the traditional modeling of stationary textures, while we strive to unify authenticity, diversity, and scalability with a neural representation to overcome limitations that existed in the traditional approaches. 

% We refer the readers to the comprehensive surveys~\cite{wei2009state,raad2018survey,pietroni2010solid} for a more detailed coverage of the traditional methods.

% How does a neural network learn the stylish representation of a texture without simply reconstructing it? Two key milestones are marked in the advances. The first is the use of Gram matrices~\cite{gatys2015texture, johnson2016perceptual, ulyanov2016texture} in a feed-forward synthesis network based on differences between high-level features, which is known as the perceptual loss or content loss. The correlation definition allows high-quality textures to be synthesized while only resembling the input image stylishly. However, feed-forward network is limited in its ability to generate various samples for an input exemplar, albeit attempt to encourage diversity~\cite{li2017diversified}. 

Compared to earlier parametric models, the artificial neural network is powerful in its generalizability in pattern recognition~\cite{hansen1990neural}. Thus neural synthesis of texture images has marked the advances in recent endeavors. How does a neural network learn the stylish representation of a texture without simply reconstructing it? A milestone that unifies both synthesized quality and sampling power adversarially trains a generator network and a discriminator network to learn the mapping from a latent space to the texture distribution~\cite{goodfellow2014generative}. We are particularly interested in the generative adversarial networks (GANs) that models the inner statistics of the exemplars. This is marked by a patch-based approach that represents an image as a collection of smaller images: \cite{li2016precomputed, jetchev2016texture} formalizes patch-based synthesis in the GAN setting with the concept of Markovian and spatial GAN. \cite{bergmann2017learning} motivated us with its periodic design in latent codes, which effectively captures stationary patterns from an input exemplar. \cite{zhou2018non} can be seen as a complement to our design by focusing on addressing non-stationary texture synthesis by learning to expand an image patch. In addition, \cite{shaham2019singan, shocher2019ingan} present multi-scale, global-focused approaches to effectively recreate natural images. While the aforementioned approaches all utilize convolutional designs, our work extends texture synthesis to the continuous domain with an implicit representation of textures, as we argue that such representation provides a more natural and efficient way to synthesize stationary patterns. 

The synthesis of 3D shapes is of particular interest in computer graphics and thus has a rich history. To name a few, this includes volumetric field design~\cite{zhang2006vector, palacios2016tensor, chen2016synthesis}, procedural generation~\cite{ijiri2008example, merrell2007example} and 3D texturing~\cite{zhou2006mesh,bhat2004geometric}. However, very few works have considered the synthesis of 3D patterns with neural networks, with the exception of~\cite{henzler2020learning,portenier2020gramgan}, which explores the generation of 3D solid textures.
\vspace{-6mm}
\paragraph{Implicit Network}
Implicit network refers to multilayer perceptron that learns a continuous function in real coordinate space. In particular, the implicit network is mainly utilized in the reconstruction of 3D shapes~\cite{park2019deepsdf, saito2019pifu, liu2019learning,sitzmann2020implicit}, where shapes are represented by a distance function, or a radiance field~\cite{mildenhall2020nerf,yu2021pixelnerf}. We are motivated by the signed distance representation of shapes and the Fourier encoding explored in~\cite{mildenhall2020nerf,tancik2020fourier} in our design of the implicit networks, and our work adopts these features to a generative setting where novel 3D patterns are synthesized.

\section{Method}
\label{sec:method}

Our method is best introduced by expanding from the Wasserstein GAN value function~\cite{gulrajani2017improved} constructed as the \textit{Earth-Mover} distance between the real data distribution $\distribution_{data}$ and the generator distribution $\distribution_g$: 

\begin{equation}
    \underset{G}{\min} \: \underset{D}{\max} \: \mathbb{E}_{\realx \sim \distribution_{data}} [\log (D(\realx))] - \mathbb{E}_{\bm{z} \sim \distribution_Z} [\log (D(G(\bm{z}))].
    \label{equ:wgan}
\end{equation}
Our first change of the above objective is to draw real-valued coordinates $\coords \in \mathbb{R}^k$ ($k=2$ for image and $k=3$ for volume) from a distribution $\{ s\coords \: \vert \: \coords \sim \distribution_{\coords} \}$ as input to the generator, where $s$ is a constant scalar. This underlies the implicit nature of the generative network, as $G$ learns a mapping from the real coordinate space in a stochastic process. Instead of being sampled from a prior distribution $\distribution_Z$, latent variables are drawn from a random field $f_z(\coords)$ defined on the real coordinate space. $G$ is therefore a function of both the coordinates $\coords \in \mathbb{R}^k$ and the latent variables $f_z(\coords) \in \mathbb{R}^d$. This updates Eq.\ref{equ:wgan} to

\begin{equation}
\begin{split}
    \underset{G}{\min} \: \underset{D}{\max} \: & \mathbb{E}_{\realx \sim \distribution_{data}} [\log (D(\realx))] - \\ & \mathbb{E}_{\coords \sim \distribution_c, \bm{z} \sim \distribution_{f_z(\coords)}} [\log (D(G(\bm{z},\coords))].
    \label{equ:wgan2}
\end{split}
\end{equation}
In the implementation, our goal is to synthesize color, distance function, normal vectors, etc. that are defined in a grid-like structure $X : \mathbb{R}^{H \times W \times C}$ or $X : \mathbb{R}^{H \times W \times D \times C}$ and thus we also sample grid coordinate input $C : \mathbb{R}^{H \times W \times 2}$ or $C : \mathbb{R}^{H \times W \times D \times 3}$ whose center is drawn from a uniform distribution $U(-s,s)$. The randomness to the grid center position is critical in encouraging smooth and seamless transition between blocks of patterns as it models the distribution of randomly sampled patches from the input pattern. 

In the following sections, key modules are discussed in detail: 1) a deformable periodic encoding of the coordinates to model stationary patterns; 2) implementation of the latent random field; 3) a conditional variance of our objective for the synthesis of directional exemplar and controllability of the synthesized patterns, and 4) the overall network structure that completes our pattern synthesis framework.

\subsection{Periodic Encoding}
\label{sec:periodic}

As discussed in the introduction, it is critical to represent stationary patterns from the input by a repeated structure that avoids simply reconstructing the original exemplar. Simply mapping spatial coordinates to the visual pattern does not satisfy this requirement: since each coordinate is unique in the real-value space, the network would learn to overfit the coordinates to their associated positions in the exemplar and therefore fail to capture a repeatable structure. 

The benefits of a periodic encoding to the coordinates are two-fold: Firstly, it disentangles patch-level appearance from their specific position in the exemplar, which allows the pattern synthesized by the generator to be shift-invariant. Secondly, recent advances in implicit network~\cite{mildenhall2020nerf, tancik2020fourier, sitzmann2020implicit} have found a Fourier feature mapping with a set of sinusoids effective in learning high-frequency signals by addressing the ``spectral bias"\cite{basri2020frequency} inherent to MLPs. In our work, we use the following periodic mapping for the input coordinates:

\begin{equation}
\begin{split}
    \encodingFunc(\coords) = [ \cos(2^0\pi a \coords), \sin(2^0\pi a \coords), &\\
    \cdot \cdot \cdot,  \cos(2^i\pi a \coords), \sin(2^i\pi a \coords)],
    \label{equ:encoding}
\end{split}
\end{equation}
where the learnable parameter $a \in \mathbb{R}^k$ determines the period of the encoding. This design allows the network to learn to match the period of the encoding to the repeatable structure of the exemplar. It thus provides robustness to the scale of the patches sampled in training as such scale no longer dictates the period of the synthesized pattern.

\subsection{Latent Random Field}
\label{sec:latent}

In a generative framework, latent noise sampled from a prior distribution model the variation in observations. A noise function that is smoothly defined in the real coordinate space encourages smooth transition between the synthesized patches. In a 2D example, we start with a discrete random field that maps a uniform grid of coordinates to random variables $\{f_z(\coords) \: \vert \: \coords : \mathbb{R}^{H \times W \times 2}\}$. Then the discrete random field is smoothly interpolated to form a smooth latent field (see the visualization in Figure~\ref{fig:network}). In our implementation, we used an exponential interpolation:

% \begin{equation*}
% f(\bm{x}) = \sum_{i=1}^{K} w_i(\bm{x}) f_z(\coords_i)
% \end{equation*}
% \begin{equation}
% w_i(\bm{x}) = \frac{\mathrm{e}^{\frac{||\bm{x} - \coords_i||_2}{\sigma} }}{\sum_{j=1}^{K}\mathrm{e}^{\frac{||\bm{x} - \coords_j||_2}{\sigma}}} 
% \end{equation},

\begin{equation}
f(\bm{x}) = \sum_{i=1}^{K} w_i(\bm{x}) f_z(\coords_i), \: w_i(\bm{x}) = \frac{\mathrm{e}^{\frac{||\bm{x} - \coords_i||_2}{\sigma} }}{\sum_{j=1}^{K}\mathrm{e}^{\frac{||\bm{x} - \coords_j||_2}{\sigma}}},
\end{equation}
where the latent code at spatial position $x$ is interpolated from $K=4$ latent vectors defined at the grid corners. In implementation, the discrete grid used to define the random field has a spacing of 1. To match the extent of a latent factor with the learned period $a$, we simply scale the uniform grid of the discrete random field accordingly.

\subsection{Conditional IPFN}
\label{sec:conditional}

\begin{figure*}[t!]
 \begin{center}
  \centering
  \includegraphics[width=0.85\linewidth]{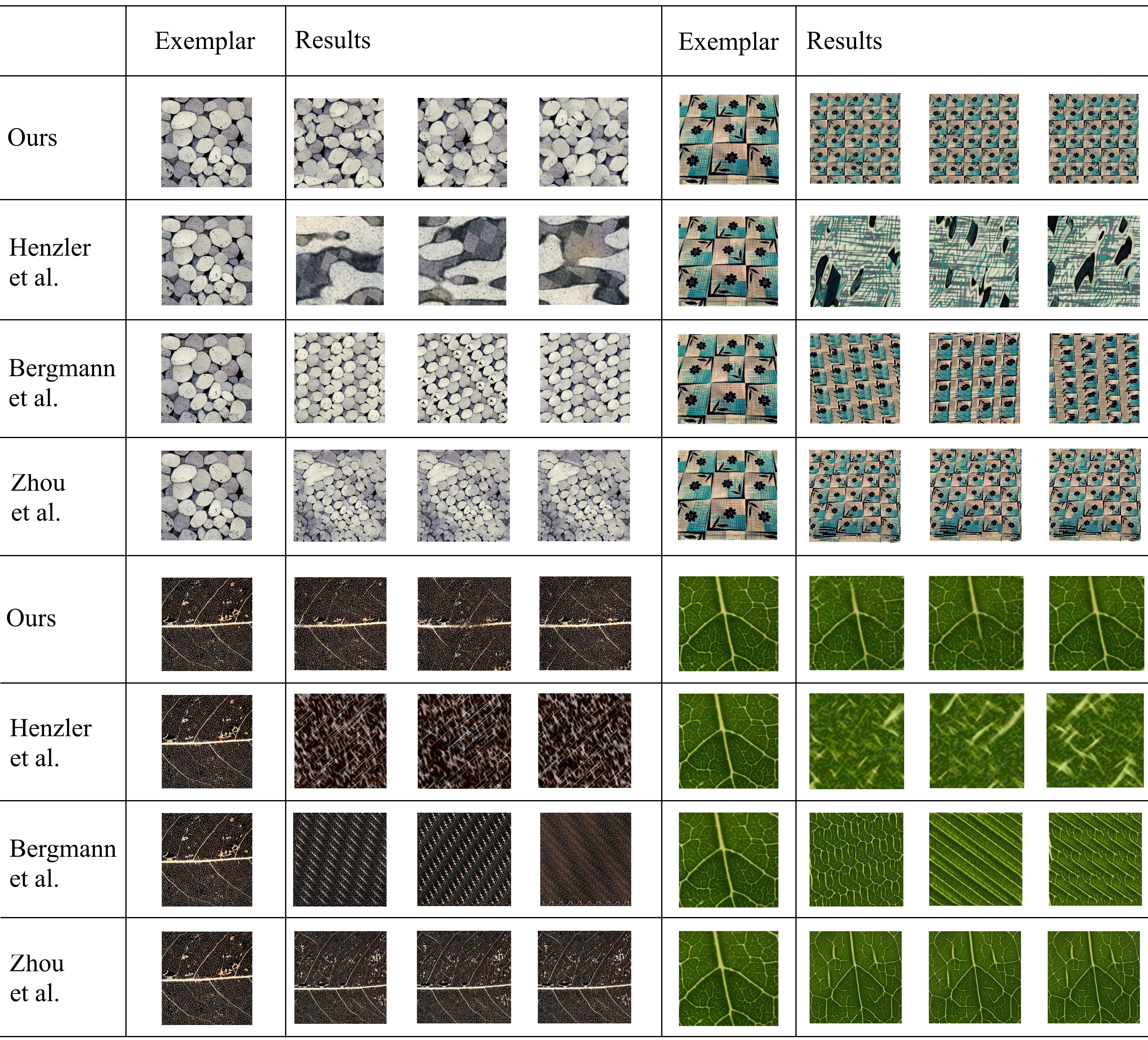}
 \end{center}
 \caption{Main results for 2D texture synthesis with comparisons to Henzler et al.~\cite{henzler2020learning}, Bergmann et al.~\cite{bergmann2017learning}, and Zhou et al.~\cite{zhou2018non} on synthesizing two stationary patterns (top four rows) and two directional patterns (bottom four rows).}{}
 \label{fig:texture2d}
\end{figure*}

Extending our GAN objective to be conditional enables many practical applications. Assume each input patch is paired with a guidance factor $\guidance$, the conditional objective is simply an extension:

\begin{equation}
\begin{split}
    \underset{G}{\min} \: \underset{D}{\max} \: & \mathbb{E}_{\realx \sim \distribution_{data}} [\log (D(\realx \vert \guidance))] - \\ & \mathbb{E}_{\coords \sim \distribution_c, \bm{z} \sim \distribution_{f_z(\coords)}} [\log (D(G(\bm{z},\coords \vert \guidance))].
    \label{equ:wgan3}
\end{split}
\end{equation}
Here we outline two applications in pattern synthesis using the conditional formulation:

\begin{itemize}
    \item \textbf{Synthesis of directional pattern}: Many natural patterns have a directional distribution that is oftentimes considered non-stationary. A typical example is a leaf texture - a midrib defines the major direction that separates the blade regions by half (see Figure~\ref{fig:texture2d}). A conditional extension of our model is able to model patch distribution along a specified direction. For simplicity, we present a 2D formulation that can be easily extended to 3D. With a user-defined implicit 2D line equation $ax + by + c = 0$, the guidance factor is defined as $\guidance(x,y) = ax + by + c$. Pixel coordinates $(p_x,p_y)$ from an input texture image with width $w$ and height $h$ are transformed as $(x,y) = (\frac{2p_x}{w} - 1,\frac{2p_y}{h} - 1)$ to be normalized to the value range of $[-1,1]$. In our experiments we have found it sufficient to condition our model on the horizontal ($y=0$) and vertical direction ($x=0$) for the evaluated exemplars. 
    \item \textbf{Controlling synthesis of 3D shapes}: In the modeling of geometric patterns, it is often desirable for the synthesis algorithm to provide explicit control of certain geometric properties such as density and orientation. These geometric properties can be calculated from the exemplar shape. Let $g(x)$ be a shape operator that defines the geometric property of interest, our conditional model trained with the sample pair $(x,g(x))$ then learns a probabilistic mapping from a guidance vector field to the target 3D shape. An intuitive example of this application can be found in Section~\ref{sec:foam_control}. 
\end{itemize}

\subsection{Network Structure}
\label{sec:network}
\begin{figure}[t]
 \begin{center}
  \centering
  \includegraphics[width=1\linewidth]{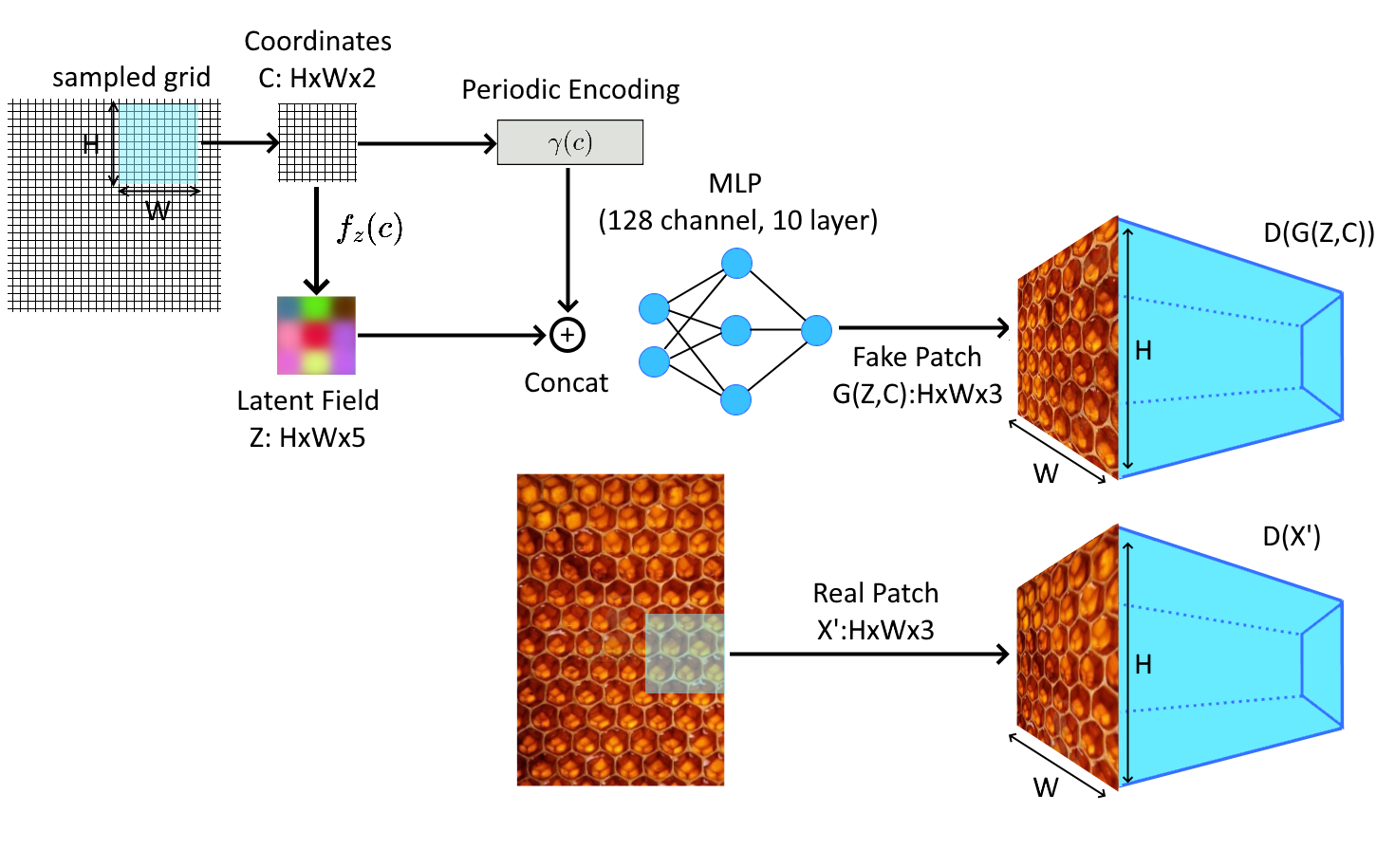}
 \end{center}
 \vspace{-6mm}
 \caption{Overview of our network architecture discussed in Section~\ref{sec:network}.}
 \label{fig:network}
  \vspace{-6mm}
\end{figure}

The overall structure of IPFN is visualized in Figure~\ref{fig:network}. The Generator Network $G$ is a 10-layer MLP with ReLU activations between layers and a sigmoid function in the end. A grid of coordinates is sampled based on a randomly shifted center. The coordinates are then passed to two separate branches: 1) the periodic encoder, and 2) a projection on the latent field to obtain a 5-dim latent vector for each coordinate. The latent codes and the periodically encoded coordinates are then concatenated as input to the generator mlp, which outputs the fake sample. The discriminator $D$ discriminates between the generated samples and randomly cropped patches from the real input. We implement $D$ by following the DCGAN architecture~\cite{radford2015unsupervised} with a stride of 2. For discriminating 3D volumes, 2D convolution layers are replaced by 3D convolution. 

%  \textbf{b.} Terrain synthesis from 26 rendered terrain views. \textbf{c.} Close-up comparisons on the synthesized honeycomb textures.

\section{Experiments}
\label{sec:result}

We hypothesize that our approach is most suitable for synthesizing texture patterns and 3D shapes with repeating structures and local variations in appearance. To this end, we demonstrate our main results by applying IPFN to the synthesis of 2D texture images and 3D structured shapes. In addition, IPFN is adapted to two applications in 3D texturing and shape manipulation. To evaluate the effectiveness of the proposed techniques, we have also conducted an ablation study where several key designs are altered. 

\textbf{Evaluation metric} While the quality for pattern synthesis is not easily quantifiable, human eyes usually provide a reasonable qualitative assessment for whether the synthesized patterns capture the aesthetics and structure of the exemplar. In our evaluation, we present comparisons of visual results that are self-manifesting, since the synthesized patterns bear obvious characteristics of the underlying designs of the synthesizer. In addition, we have provided quantifiable metrics in terms of Single Image Frech\'et Inception Distance and inference time and memory.

% The results will be presented with an emphasis on the three aforementioned factors: 1) \textit{authenticity}, which represents the visual similarity between the patterns and the exemplar. 2) \textit{diversity}, which can be measured both across and within different samples, and 3) \textit{scalability}, which is demonstrated by the capability of IPFN to efficiently synthesize large-scale patterns in both 2D and 3D.

\textbf{Implementation details} For all of our experiments, the network is optimized under WGAN loss with gradient penalty~\cite{gulrajani2017improved}. Adam optimizer~\cite{kingma2014adam} is used with a learning rate of $1e-4$ for both the discriminator D and the generator G. In each iteration, both D and G are updated for 5 steps sequentially. Input images and volumes are randomly cropped to a smaller-scale patch. For positional encoding, we choose a bandwidth $i=5$ as a wider kernel tends to produce sinusoidal artifacts, whereas a narrower kernel produces blurry results. The input coordinates are randomly shifted by an offset in the range $[-4,4]$ to accommodate for the chance that the network may learn an increased period for the periodic encoding. Accordingly, noises are interpolated from a $5 \times 5$ grid ($5^3$ for 3D volume) discrete random field, where the point locations are ranged between $[-5,5]$. A single-exemplar experiment is typically trained for 12,500 iterations with a batch size of 8 and runs on a single Nvidia GTX 1080 GPU, which takes about 6-9 hours to complete. \textbf{Inference Time}: IPFN only requires 24 milliseconds to generate a $1024 \times 1024$ image. 3D volumes are generated iteratively and a large-scale $512^3$ volume takes only 22.9 seconds to be generated. Source code will be made publicly available upon acceptance.

\subsection{Texture Pattern Synthesis}
\label{sec:exp_texture}

Image sources selected from the Oxford Describable Textures Dataset (DTD)~\cite{cimpoi2014describing} are shown in Figure~\ref{fig:texture2d}. Specifically, we selected two exemplars with stationary patterns (top 4 rows in Figure~\ref{fig:texture2d}) and two exemplars with directional patterns (bottom 4 rows in Figure~\ref{fig:texture2d}) to demonstrate that IPFN synthesizes visually similar patterns in both cases. During training, images were randomly cropped into patches of size $128 \times 128$. During inference, the synthesized images were scaled up four times to a size of $512 \times 512$. Our results are compared to the three most relevant baseline generative methods that synthesize texture patterns from a single image:
\begin{itemize}
    \item Henzler et al.~\cite{henzler2020learning}: A method that similarly utilizes implicit network and smooth noise field for texture synthesis. The synthesized results were obtained from running the officially released code.
    \item Bergmann et al.~\cite{bergmann2017learning}: A convolutional method that combines noise with periodic signals to synthesize stationary pattern. The synthesized results were obtained from running the officially released code.
    \item Zhou et al.~\cite{zhou2018non}: A convolutional image expansion approach targeted for non-stationary texture synthesis. Since \cite{zhou2018non} expands from an image input deterministically and does not utilize latent code, only one synthesized result is shown per row. The synthesized results were obtained from the authors.
\end{itemize}

Visual inspection is sufficient to show that IPFN provides promising results. When compared to~\cite{henzler2020learning}, IPFN synthesized results with obvious structures as noise is not directly mapped to the output. While~\cite{bergmann2017learning} synthesizes periodic samples that display diversity across samples and similarity to the stationary exemplars, their synthesized patterns lack variation within the image. In comparison, our synthesized patterns show a higher level of local variations and adapt well to the directional cases. \cite{zhou2018non} has provided the most visually authentic results among the baselines. However, in the stationary cases, radial distortion is noticeable near the boundaries of its synthesized images. Moreover, without requiring image input, IPFN provides a more direct approach to synthesizing diversified samples from random noise. 

% We have additionally provided large-scale honeycomb results in Figure~\ref{fig:texture2d}.c to amplify the comparison.

% In addition, we also show in Figure~\ref{fig:texture2d}.b synthesis results from multiple rendered images of a terrain scene. The terrain scene is selected from the BlendedMVS dataset~\cite{yao2020blendedmvs}, where we render 26 top-down views based on the reconstructed mesh. IPFN is able to combine elements present in the scene (cottage, snowfield, split-level house) and synthesize them in various arrangements. This demonstrates that IPFN has a potential application in large-scale terrain synthesis. 

\begin{table}[b]
\small
\centering
\begin{tabular}{|c|c|c|c|c|}
\hline
  & honey &  crosshatch & rock & leaf \\ \hline
Henzler~\cite{henzler2020learning} &    332.66  &      310.49   &   351.23  &  225.11            \\ \hline
Bergmann~\cite{bergmann2017learning} &    62.75  &      177.88   &   120.64 &  164.37            \\ \hline
Zhou~\cite{zhou2018non} &   14.54   &  154.63       &   118.29  &  \textbf{38.13}            \\ \hline
Ours &    \textbf{10.15}  & \textbf{130.83}  &  \textbf{113.81}  &  103.6            \\ \hline
\end{tabular}
\caption{SIFID scores between the exemplars and the generated patterns from ours and different baselines. }
\label{tab:sifid}
\end{table}

\textbf{Single Image Frech\'et Inception Distance (SIFID)}  SIFID introduced in~\cite{shaham2019singan} is a metric commonly used to assess the realism of generated images. 
For the computation of SIFID, we have used a patch size of $128 \times 128$ in all experiments, where the synthesized patterns have the same resolution as the original exemplars. Table~\ref{tab:sifid} shows the SIFID comparisons between ours and the baselines in various categories of exemplars. For Zhou et al.~\cite{zhou2018non}, only the generated (expanded) portion of the images were used. The results show that our method can generate results that better resemble the distribution of the real texture in the stationary categories (honey, crosshatch, rock) as our generated patterns receive lower SIFID scores. For the leaf category, a typical directional pattern, Zhou et al.~\cite{zhou2018non} achieves the best performance as its method specifically targets non-stationary expansions, while our method still performs better than other baselines that have not taken into consideration the synthesis of non-stationary patterns. 

% The \textbf{lower} the SIFID score is, the \textbf{closer} the generated patches resemble the distribution of the real texture. 

% 

% \begin{table}[h]
% \small
% \centering
% \begin{tabular}{|c|c|c|c|}
% \hline
%   & honey &  crosshatch & rock \\ \hline
% Henzler [12] &    332.66  &      310.49   &   351.23              \\ \hline
% Bergmann [2] &    62.75  &      177.88   &   120.64            \\ \hline
% Zhou [43] &   14.54   &  154.63       &   118.29             \\ \hline
% Ours &    \textbf{10.15}  & \textbf{130.83}  &  \textbf{113.81}            \\ \hline
% \end{tabular}
% \caption{blablablblbalbala}
% \end{table}

\begin{table}[t]
\small
\centering
\begin{tabularx}{8.6cm}{|X|c|c|c|c|}
\hline
 Time (ms) / memory (GB) & $128^2$ & $256^2$ & $512^2$ & $1024^2$ \\ \hline
Henzler~\cite{henzler2020learning} &    218/1.38  &      278/1.62   &   328/2.72  &  458/6.45            \\ \hline
Bergmann~\cite{bergmann2017learning} &   \textbf{7}/2.37  & 13/5.79   &   42/19.68  &  115/31.88            \\ \hline
Zhou~\cite{zhou2018non} &  356/1.20    &  349/1.34       &   510/2.00  &  612/4.66            \\ \hline
Ours &    8/\textbf{0.76}  &      \textbf{11}/\textbf{0.85}   &   \textbf{15}/\textbf{1.23}  &  \textbf{24}/\textbf{2.81}            \\ \hline
\end{tabularx}
\caption{Comparisons of inference time and inference memory consumption, measured in milliseconds (ms) / gigabytes (GB), when patterns of increasing size (top row) are generated. }
\label{tab:itfn_time}
\vspace{-8pt}
\end{table}

\textbf{Inference Time and Memory} Table~\ref{tab:itfn_time} measures the inference time and memory consumption of our network compared to the baselines when generating image at different sizes. Our implicit formulation is shown to be significantly more efficient in both time and space without the needs to rely on computation of pseudo-random noise (\cite{henzler2020learning}) or convolutional operations (\cite{bergmann2017learning, zhou2018non}). This validates our claim on the scalability of our design.

\subsection{Ablation Study}
\label{sec:exp_ablation}

\begin{figure}[h]
 \begin{center}
  \includegraphics[width=0.7\linewidth]{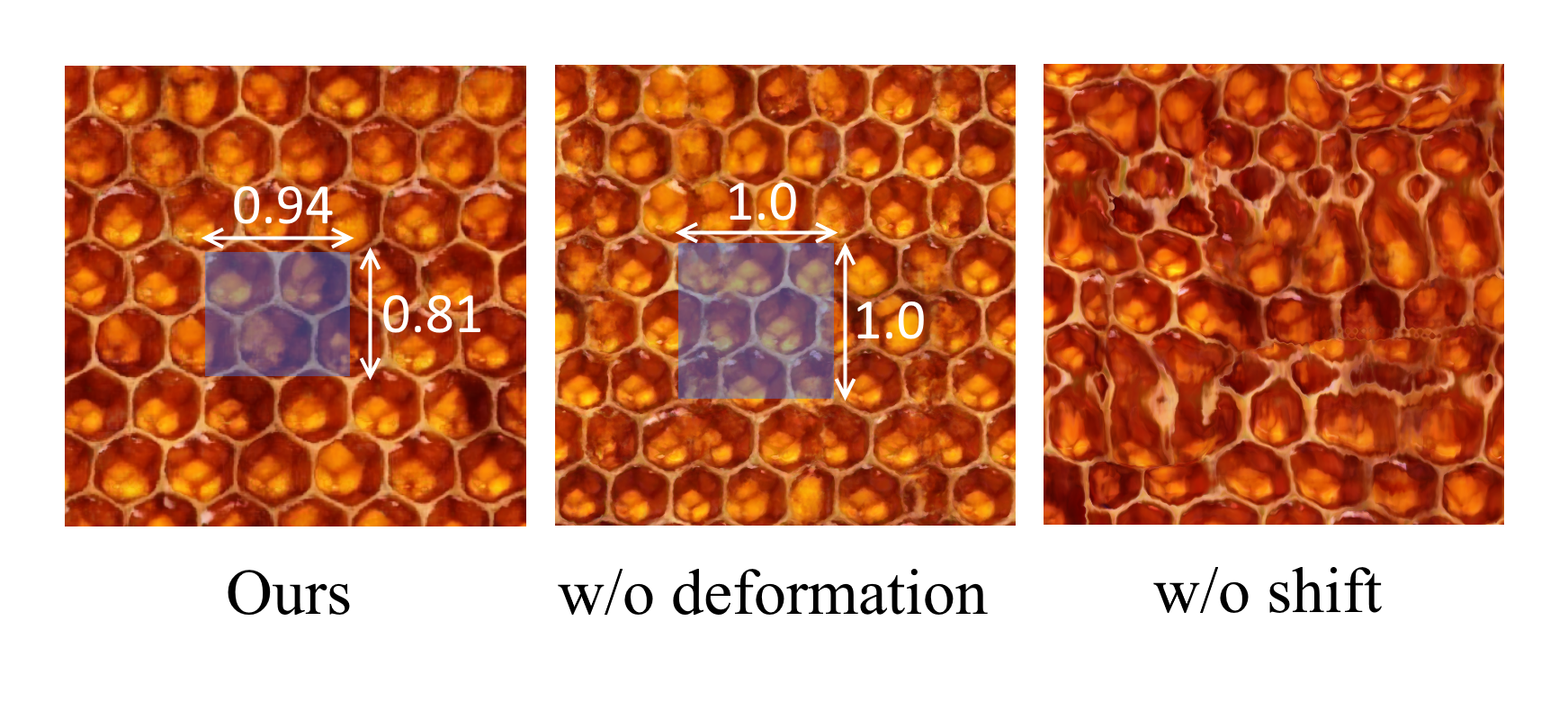}
 \end{center}
 \vspace{-8mm}
\caption{Synthesized honeycomb textures for the ablation study. The blue boxes represent the learned scale of periodic encoding in ours, where in w/o deformation, the period is default to 1, which does not match with the repeating structure of the honeycomb pattern and results in visual artifact.}

% The designs for the compared models are described in Section.~\ref{sec:exp_ablation}. 
% The noticeable differences in visual quality verify the effectiveness of our design components to capture stationary patterns present in the exemplar image.}
\centering
 \label{fig:ablation}
\vspace{-4mm}
\end{figure}

To validate our design choices, we have conducted an ablation study by removing two designs that we consider critical for our network to be effective. The comparison results are shown in Figure~\ref{fig:ablation}. \textit{w/o deformation} is a network model that encodes input coordinates without the learnable parameters $a$ described in Section~\ref{sec:periodic}. \textit{w/o shift} is a model that is trained without randomly shifting the input coordinates. The resulted patterns are indicative of the effects of these designs: when coordinates are encoded at a fixed scale, the \textit{w/o deformation} model generates hexagons that are seemingly glued together as the presumed scale does not match with the actual period of the repeating structure. The \textit{w/o shift} model synthesized distorted patterns as we speculate that, without the random sampling of the input coordinates, the network faces difficulty in matching the patch-based priors of the image patches.

\subsection{Volumetric Shape Synthesis}
\label{sec:exp_volume}

\begin{figure}[t]
 \begin{center}
  \centering
  \includegraphics[width=1.0\linewidth]{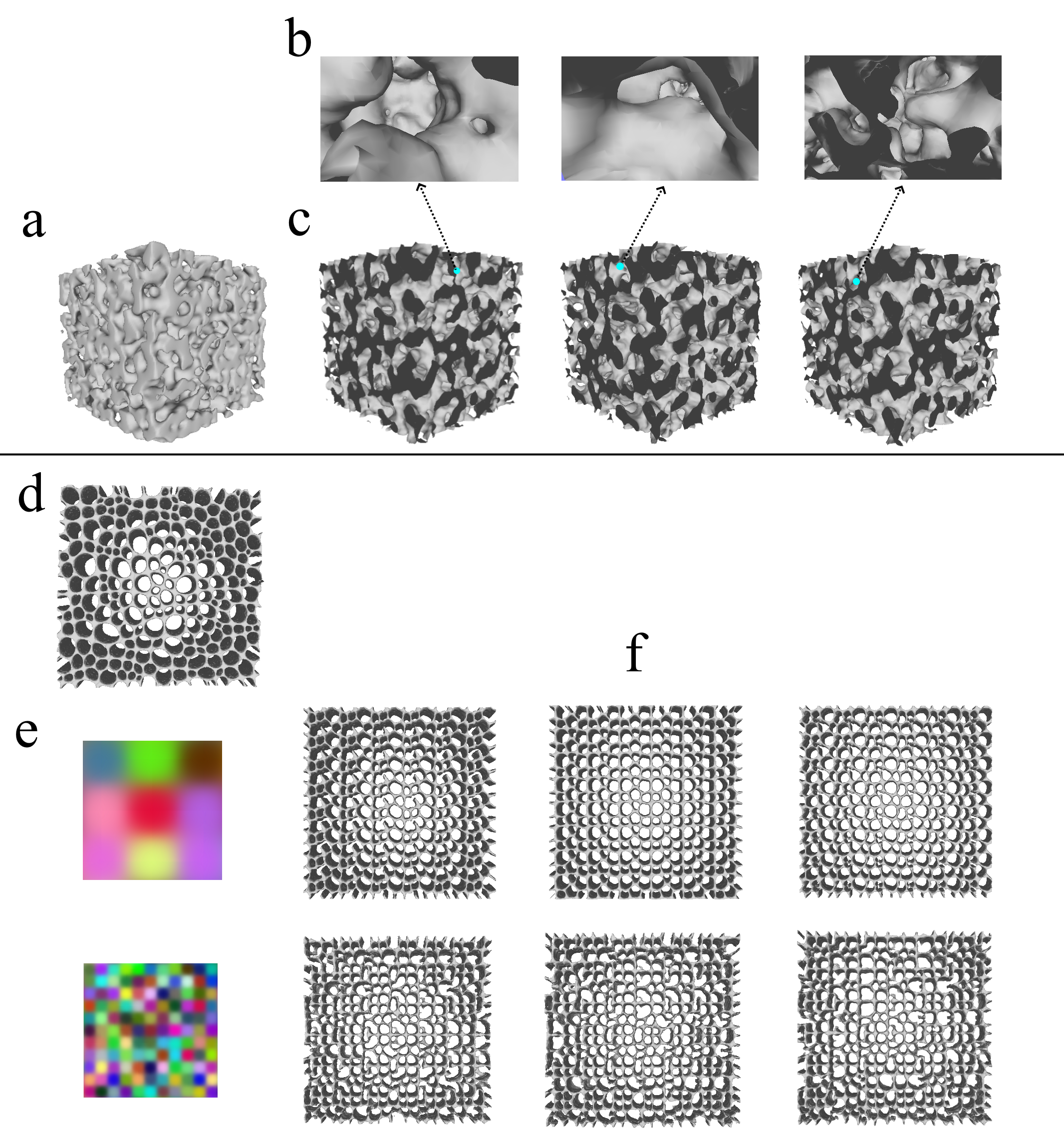}
 \end{center}
 \vspace{-4mm}
 \caption{Main results for 3D volume synthesis. \textbf{a.} Exemplar porous structure. \textbf{b.} Synthesized structure models interior tunnels. \textbf{c.} Global views of synthesized porous structures. \textbf{c.} Exemplar foam structure. \textbf{e.} Two scales of noise fields for the foam structure synthesis. \textbf{f.} Synthesized foam structures. Larger scale of the noise field leads to more isotropic foam structures.}{}
 \label{fig:volume}
 \vspace{-6mm}
\end{figure}
\begin{figure}[!t]
 \begin{center}
  \centering
  \includegraphics[width=0.8\linewidth]{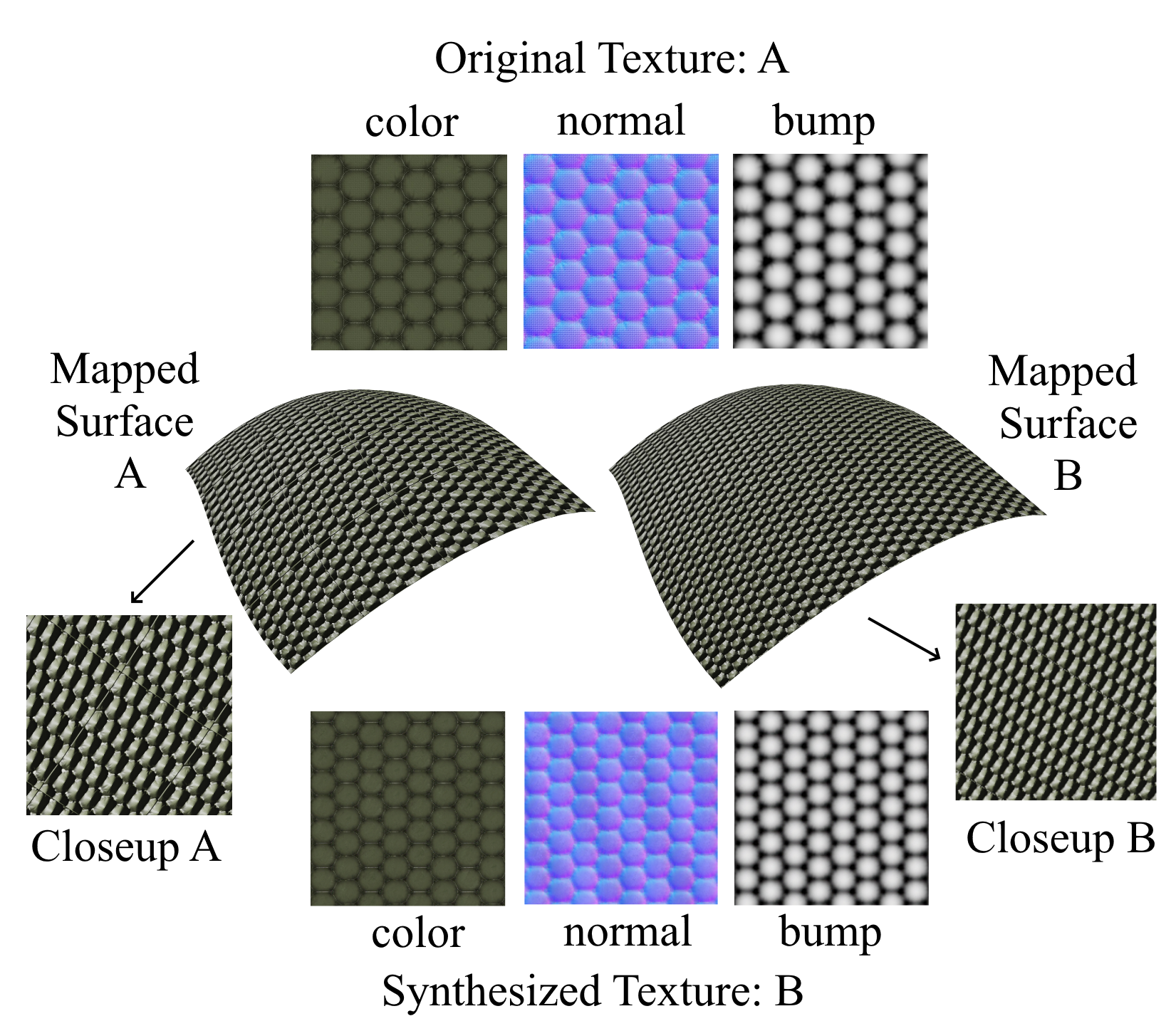}
  \vspace{-8pt}
 \end{center}
 \caption{IPFN learns multi-channel textures that are applicable to seamless 3D texturing. The original 3D texture in this example is not symmetric and therefore visible seams can be found on the texture-mapped surface and in the closeup view (A in figure). As synthesized patterns learnt from this exemplar can be tiled in any direction, the mapped surface (B in surface) is seamless.}
 \label{fig:seamless}
 \vspace{-8pt}
\end{figure}

For the evaluation on volumetric shape synthesis, We have obtained two 3D models from \textit{turbosquid.com}: a porous structure (Figure~\ref{fig:volume}.a) and a foam structure ((Figure~\ref{fig:volume}.d). The 3D meshes are preprocessed into signed distance fields by uniformly sampling points in a volumetric grid. For the porous structure, we have sampled $256 \times 256 \times 256$ points and extracted $64 \times 64 \times 64$ patches during training. For the foam structure, we have sampled $200 \times 200 \times 128$ points and extracted $32 \times 32 \times 32$ patches during training. During inference, porous structures are synthesized at their original resolution, while we scale the synthesized foam structures to be twice as large as the original shape in the XY direction. Figure~\ref{fig:volume}.c and Figure~\ref{fig:volume}.f show the synthesized shapes. For the porous structures, both outer structures and interior structures are learned (see Figure~\ref{fig:volume}.b for zoom-in interior views) and the structures are diversified both across and within samples. For the foam structures, we have shown different results by varying the extent of the latent random field (see Figure~\ref{fig:volume}.e). A larger-scale random field encourages the synthesizer to generate globally varied structures, whereas a smaller scale produces locally anisotropic structures.

\subsection{Application: Seamless 3D Texturing}
\label{sec:3dtex}

Due to the periodic nature of the synthesized patterns, noise manipulation allows IPFN to create textures that are mirror-symmetric. This property provides an immediate application to seamless 3D texture mapping: in Figure~\ref{fig:seamless}, the original 9-channel texture, composed of color, normal, and bump maps, is tiled and directly mapped to a planar surface. Due to discrepancies on the edges, as the textures are wrapped to create the tiled patterns, the mapped surfaces show visible seams. We recreate this texture through our network under a constant latent vector (Figure~\ref{fig:seamless} B). When repeatedly mapped to the surface, the symmetric texture is seamless while faithfully reflecting the appearance and structure of the original texture. 

\begin{figure}[t]
 \begin{center}
  \centering
  \includegraphics[width=0.8\linewidth]{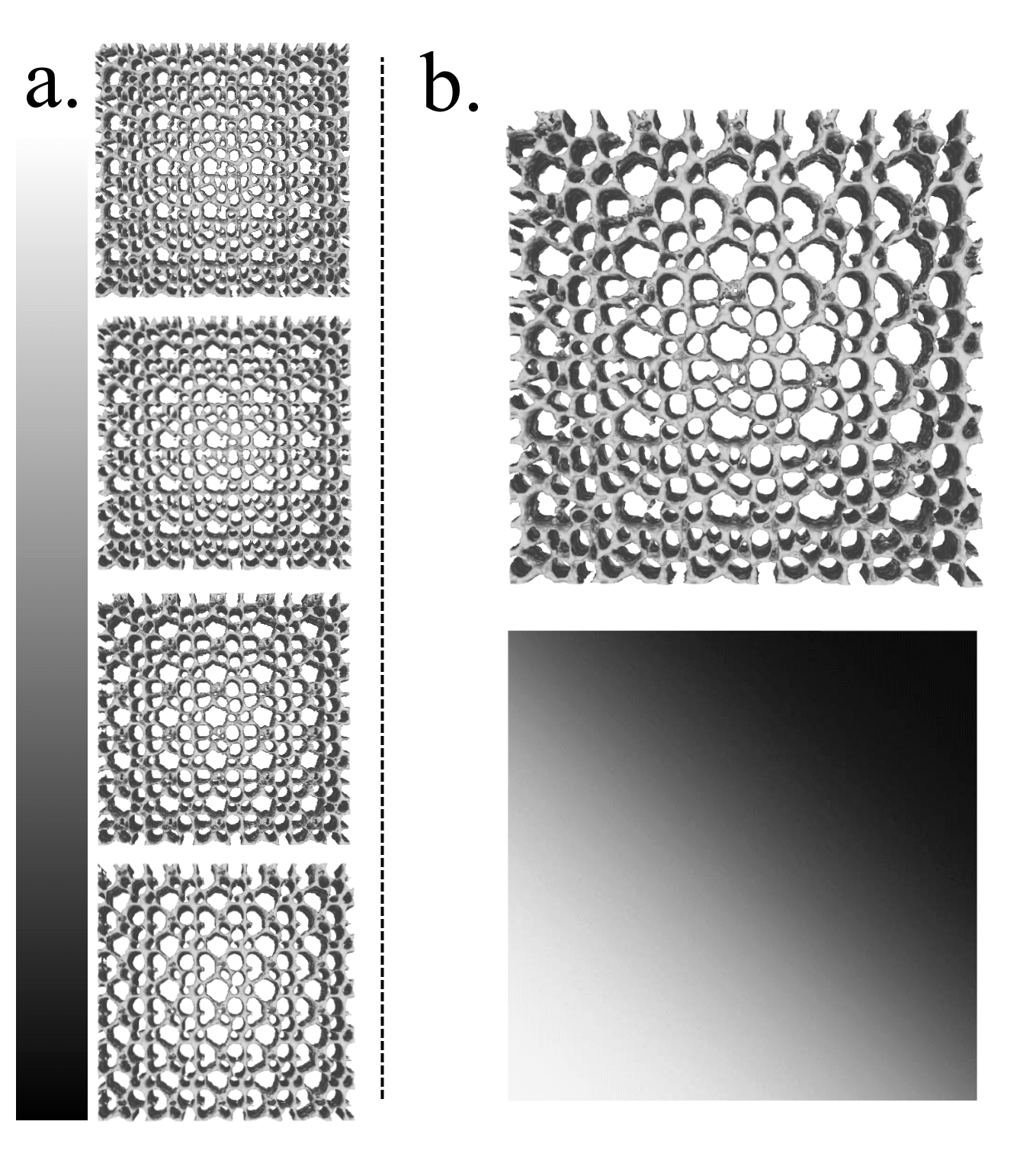}
 \end{center}
 \vspace{-5mm}
 \caption{Synthesized foam structures with controllable density. \textbf{a}. The grey scale bar controls the synthesized structure from the highest density (white) to the lowest (black). \textbf{b}. Smooth interpolation of the guidance factor allows us to synthesize a foam structure with smoothly changing densities.}
 \label{fig:volumecontrol}
 \vspace{-5mm}
\end{figure}

\subsection{Application: 3D Foam with controllable density}
\label{sec:foam_control}

The original foam shape used in our experiment contains holes of various sizes, which corresponds to the density of the foam structure. This geometric property $g$ can be easily approximated with an unsigned distance field representation. For a patch $X^\prime$ of size $H^\prime \times W^\prime \times D^\prime$, we estimate its density by:
\begin{equation}
    g(X^\prime) = \frac{1}{m}\sum_{i=1}^{H^\prime}\sum_{j=1}^{W^\prime}\sum_{k=1}^{D^\prime} \vert sdf(X_{ijk}) \vert
\end{equation},
where m is a normalization factor. Figure~\ref{fig:volumecontrol} shows the synthesized foam structures by gradually increasing the density factor (Figure~\ref{fig:volumecontrol}.a) and by a linearly interpolated density map ((Figure~\ref{fig:volumecontrol}.b).

\section{Limitations and Discussions}
\label{sec:conclusion}

\begin{figure}[h]
 \begin{center}
  \centering
  \includegraphics[width=1.0\linewidth]{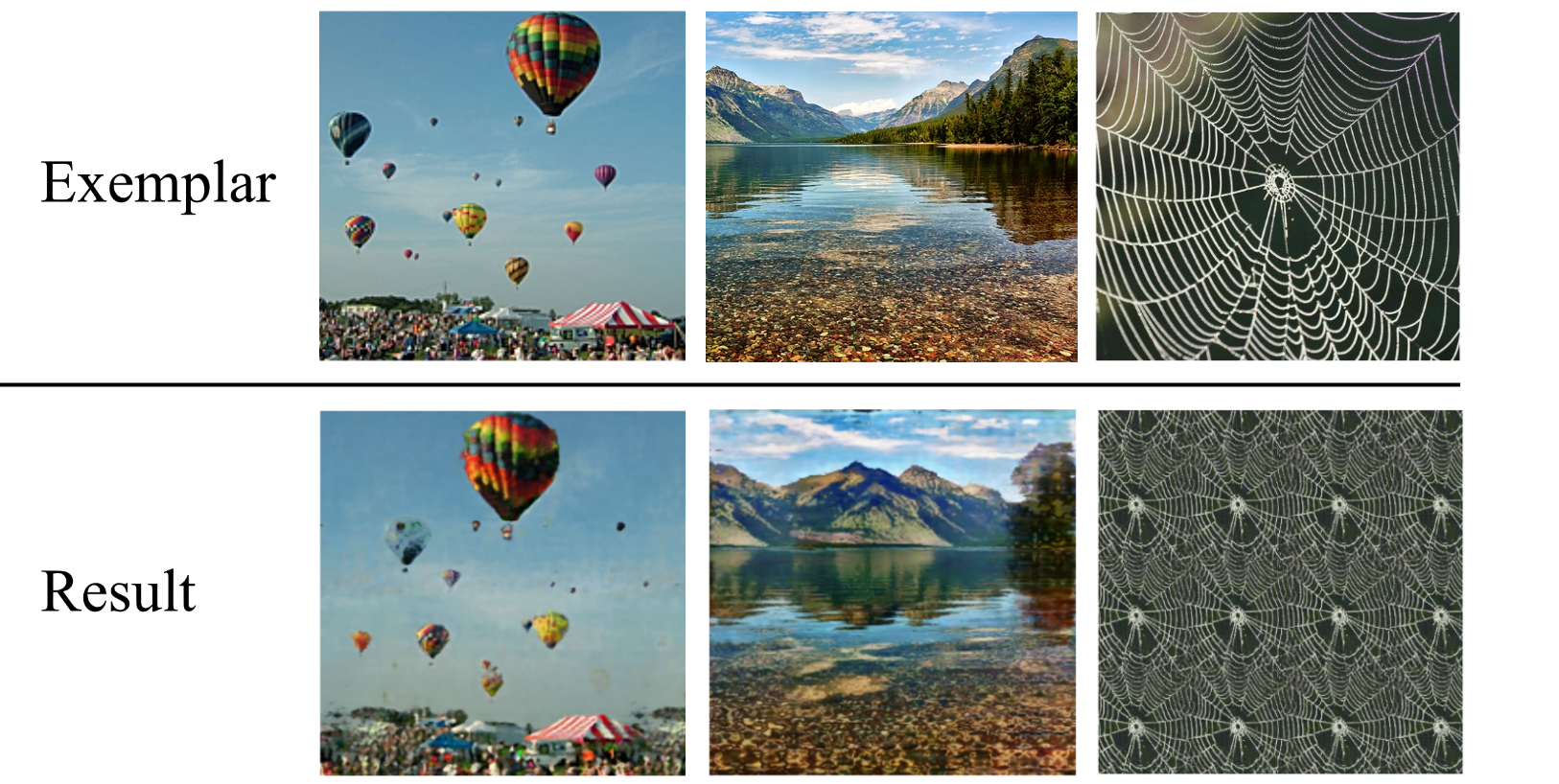}
 \end{center}
 \vspace{-4mm}
 \caption{The examples demonstrating limitations of our network}
% Details are discussed in Section.~\ref{sec:conclusion}.
 \label{fig:limitation}
 \vspace{-4mm}
\end{figure}

The main limitation of our method is its emphasis on modeling stationary patterns. While this is based on our observation that a broad range of natural patterns is stationary or directional, our method does not provide a natural way to address a class of patterns that are radial, which are exemplified by web structures and spiral patterns (see the third column in Figure~\ref{fig:limitation}). 

While our conditional formulation is in theory compatible with the synthesis of landscape images, experiments found that the quality of the synthesized landscapes are subpar - while the synthesized landscape appears globally similar to the exemplar, some local regions contain "fading-out" elements that are blended with the background (see the first two columns in Figure~\ref{fig:limitation}). We speculate that this phenomenon is due to an under-representation of these elements in the exemplar.

The above limitations have inspired us to consider many potential improvements of our methods in future works. A multi-scale synthesis approach marked in~\cite{shaham2019singan,shocher2019ingan} strikes a good balance between learning the distribution of global structure and local, high-frequency details of an image. Different geometric encoding schemes may also extend our framework to synthesize beyond stationary patterns. We believe there are still ample opportunities for the extension of our methods to a broader range of 3D applications.

% \textbf{Social Implications}: Although GAN technologies are potentially subject to malicious use in generating identity-infringing fake images or videos, we believe our specific focus on visual patterns are relatively immune to inappropriate applications. The particular patterns generated by our network (e.g. Figure~\ref{fig:volume},~\ref{fig:seamless},~\ref{fig:volumecontrol}) may trigger people with trypophobia. Therefore it is advised that a warning is included during use of our generated patterns.

\section*{Acknowledgements}
This research was sponsored by the Army Research Office and was accomplished under Cooperative Agreement Number W911NF-20-2-0053, and by the U.S. Army Research Laboratory (ARL) under contract number W911NF-14-D-0005. The views and conclusions contained in this document are those of the authors and should not be interpreted as representing the official policies, either expressed or implied, of the Army Research Office or the U.S. Government. The U.S. Government is authorized to reproduce and distribute reprints for Government purposes notwithstanding any copyright notation.
\clearpage

{\small
	\bibliographystyle{ieee_fullname}
	\bibliography{paper}
}

\end{document}

\maketitle

\section{Proofs of equivariance}

In this section, we provide proofs of \seThree{} equivariance to the convolution introduced in the main text. Recall that the \seThree{} space can be factorized into the space of 3D rotation $\{R \vert R \in \soThree{}\}$ and 3D translation $\{\translation \vert \translation \in \mathbb{R}^3\}$. A convolution operator equivariant to \seThree{} must therefore satisfy: 
\begin{equation}
\begin{split}
  \forall \rotation \in \soThree, \rotation(\SphFeat \ast \kernelFunc)(\point, \soAnc) &= (\rotation\SphFeat \ast \kernelFunc)(\point, \soAnc), \\
    \forall \translation \in \mathbb{R}^3, \translation(\SphFeat \ast \kernelFunc)(\point, \soAnc) &= (\translation\SphFeat \ast \kernelFunc)(\point, \soAnc).
  \end{split}
\end{equation}
% We prove that the proposed \convNameFull{} is equivariant to: 1) group rotation; 2) point cloud translation.

\noindent \textbf{Theorem 1.} The continuous convolution operator \begin{multline}
  (\SoFeat \ast \kernelFunc)(\point, \soAnc) \\ =
  \int_{\pointi \in \mathbb{R}^{3}} \int_{\soAncj \in \soThree} \SoFeat(\pointi, \soAncj) \kernelFunc(\soAnc^{-1}(\point - \pointi), \soAncj^{-1}\soAnc)
  \label{equ:6d_conv_continuous}
\end{multline}
is equivariant w.r.t. rotation $\rotation \in \soThree{}$ and translation $\translation \in \mathbb{R}^3$

\textit{Proof.} Firstly, we prove that Eq.(\ref{equ:6d_conv_continuous}) is equivariant to 3D rotation. For convenience of notation, let $\pointi'=\rotation^{-1}\pointi$, and $\soAncj'=\rotation^{-1}\soAncj$. 

% Due to the closure propoerty of group, 
% \begin{equation}
%     \begin{split}
%     \forall \rotation \in \soThree{}, \{g \vert g \in \soThree{}\} &= \{\rotation g \vert g \in \soThree{}\}, \\
%   \forall x \in \mathbb{R}^3, \{g \vert g \in \soThree{}\} &= \{\rotation g \vert g \in \soThree{}\},
% %   \int_{\soAncj \in \soThree}\kernelFunc(\soAnc^{-1}(\point - \pointi), \soAncj^{-1}\soAnc) \\
% %   = \int_{\soAncj \in \soThree}&\kernelFunc(\soAnc^{-1}(\point - \pointi), \soAncj^{-1}\soAnc).
%     \end{split}
% \end{equation}

% By the definition of rotation group, $\{\soAncj'\}$ is a permutation of $\{\soAncj\}$.

\begin{align*}
\hspace{-1mm}&\rotation(\SoFeat \ast \kernelFuncInter)(\point, \soAnc) = (\SoFeat \ast \kernelFuncInter)(\rotation\point, \rotation\soAnc) \\
  &= \int_{\pointi \in \mathbb{R}^{3}} \int_{\soAncj \in \soThree} \\
    &\hspace{17mm} \SoFeat(\pointi, \soAncj)  \kernelFunc((\rotation\soAnc)^{-1}(\rotation\point - \pointi), \soAncj^{-1}\rotation\soAnc) \\
    &= \int_{\pointi \in \mathbb{R}^{3}} \int_{\soAncj \in \soThree} \\
    &\hspace{17mm}\SoFeat(\pointi, \soAncj) 
    \kernelFunc(\soAnc^{-1}(\point - \rotation^{-1}\pointi), (\rotation^{-1}\soAncj)^{-1}\soAnc) \\
    \vspace{2mm}
      &= \int_{\pointi' \in \mathbb{R}^{3}} \int_{\soAncj' \in \soThree}  \\
    &\hspace{17mm}\SoFeat(\rotation\pointi', \rotation\soAncj')  \kernelFunc(\soAnc^{-1}(\point - \pointi'), \soAncj'^{-1}\soAnc) \\
  &= (\rotation \SoFeat \ast \kernelFuncInter)(\point, \soAnc).
\end{align*}

Then, we prove that Eq.(\ref{equ:6d_conv_continuous}) is equivariant to 3D translation. Let $\pointi'=\translation^{-1}\pointi$. Because $\translation(x - x_i) = x - x_i $:

\begin{align*}
\hspace{-1mm}&\translation(\SoFeat \ast \kernelFuncInter)(\point, \soAnc) = (\SoFeat \ast \kernelFuncInter)(\translation\point, \soAnc) \\
  &= \int_{\pointi \in \mathbb{R}^{3}} \int_{\soAncj \in \soThree} \\
    &\hspace{17mm} \SoFeat(\pointi, \soAncj)  \kernelFunc(\soAnc^{-1}(\translation\point - \pointi), \soAncj^{-1}\soAnc) \\
    &= \int_{\pointi \in \mathbb{R}^{3}} \int_{\soAncj \in \soThree} \\
    &\hspace{17mm}\SoFeat(\pointi, \soAncj) 
    \kernelFunc(\soAnc^{-1}\translation(\point - \translation^{-1}\pointi), \soAncj^{-1}\soAnc) \\
    &= \int_{\pointi' \in \mathbb{R}^{3}} \int_{\soAncj \in \soThree} \\
    &\hspace{17mm}\SoFeat(\translation\pointi', \soAncj) 
    \kernelFunc(\soAnc^{-1}(\point - \pointi'), \soAncj^{-1}\soAnc) \\
  &= (\translation \SoFeat \ast \kernelFuncInter)(\point, \soAnc).
\end{align*}

The continuous convolution operator is therefore \seThree{} equivariant.
Given a finite point set $\Point$ and a finite rotation group $\SoAncGroup$, the \seThree{} separable convolution consists of two discrete convolution operators: 
\begin{align}
    (\SoFeat \ast \kernelFuncInter)(\point, \soAnc) &= \sum_{\pointi \in \Point} \SoFeat(\pointi) \kernelFuncInter(\soAnc^{-1}(\point - \pointi), \soAnc)
    \label{equ:separate_inter} \\
    (\SoFeat \ast \kernelFuncIntra)(\point, \soAnc) &= \sum_{\soAncj \in \SoAncGroup} \SoFeat(\soAncj) \kernelFuncIntra(\point, \soAncj^{-1}\soAnc)
  \label{equ:separate_intra}
\end{align}

For convenience, we use an equivalent definition in the following proof:
\begin{align}
    (\SoFeat \ast \kernelFuncInter)(\point, \soAnc) &= \sum_{\pointi \in \Point} \SoFeat(\pointi, \soAnc) \kernelFuncInter(\soAnc^{-1}(\point - \pointi))
    \label{equ:separate_inter} \\
    (\SoFeat \ast \kernelFuncIntra)(\point, \soAnc) &= \sum_{\soAncj \in \SoAncGroup} \SoFeat(\point, \soAncj) \kernelFuncIntra(\soAncj^{-1}\soAnc)
  \label{equ:separate_intra}
\end{align}

\noindent \textbf{Theorem 2.} The discrete convolution operators given in Eq.(\ref{equ:separate_inter}),(\ref{equ:separate_intra}) are equivariant w.r.t. rotation $\rotation \in \SoAncGroup$ and translation $\translation \in \mathbb{R}^3$

Again, we first prove that the two operators are equivariant to 3D rotations in the rotation group $\SoAncGroup$. Following the notations used in the previous proof, let $\Point_\rotation = \{\pointi' \vert \pointi' = \rotation \point , \point \in \Point  \}$, $G_\rotation = \{ g_j' \vert g_j' = \rotation^{-1} g, g \in G \}$:
% $\soAnc' = \rotation \soAnc$ $\SoAncGroup_\rotation = \{\rotation\soAnc \vert \soAnc \in \SoAncGroup\}$. Due to the closure property of group, $\forall \rotation \in \SoAncGroup, \soAnc' = \rotation\soAnc \in \SoAncGroup$:

\vspace{-5mm}
\begin{align*}
&\rotation(\SoFeat \ast \kernelFuncInter)(\point, \soAnc) = (\SoFeat \ast \kernelFuncInter)(\rotation\point, \rotation\soAnc) \\ &= \sum_{\pointi \in \Point} \SoFeat(\pointi, \rotation\soAnc) \kernelFuncInter((\rotation\soAnc)^{-1}(\rotation\point - \pointi))
\\ &= \sum_{\pointi \in \Point} \SoFeat(\pointi, \rotation\soAnc) \kernelFuncInter(\soAnc^{-1}(\point - \rotation^{-1}\pointi))
\\ &= \sum_{\pointi' \in \Point_\rotation} \SoFeat(\rotation \pointi', \rotation\soAnc) \kernelFuncInter(\soAnc^{-1}(\point - \pointi')) \\
&= (\rotation \SoFeat \ast \kernelFuncInter)(\point, \soAnc).
\\
\\
&\rotation(\SoFeat \ast \kernelFuncIntra)(\point, \soAnc) = (\SoFeat \ast \kernelFuncIntra)(\rotation\point, \rotation\soAnc) \\ &= \sum_{\soAncj \in \SoAncGroup} \SoFeat(\rotation\point, \soAncj) \kernelFuncIntra( \soAncj^{-1}\rotation\soAnc)
\\ &= \sum_{\soAncj' \in \SoAncGroup_\rotation} \SoFeat(\rotation\point, \rotation \soAncj') \kernelFuncIntra( \soAncj'^{-1}\soAnc)
\\
&=( \rotation \SoFeat \ast \kernelFuncIntra)(\point, \soAnc).
\end{align*}

We then prove that the two operators are equivariant to 3D translation. Let $\pointi' = \translation^{-1} \pointi$:
\vspace{-1mm}
\begin{align*}
&\translation(\SoFeat \ast \kernelFuncInter)(\point, \soAnc) = (\SoFeat \ast \kernelFuncInter)(\translation\point, \soAnc) \\ 
&= \sum_{\pointi \in \Point} \SoFeat(\pointi, \soAnc) \kernelFuncInter(\soAnc^{-1}(\translation\point - \pointi))
\\&= \sum_{\pointi \in \Point} \SoFeat(\pointi, \soAnc) \kernelFuncInter(\soAnc^{-1}\translation(\point - \translation^{-1}\pointi))
\\ &= \sum_{\pointi \in \Point} \SoFeat(\pointi, \soAnc) \kernelFuncInter(\soAnc^{-1}(\point - \translation^{-1} \pointi)) \\ &= \sum_{\pointi' \in \translation^{-1}\Point} \SoFeat(\translation \pointi', \soAnc) \kernelFuncInter(\soAnc^{-1}(\point - \pointi')) \\
&= (\translation \SoFeat \ast \kernelFuncInter)(\point, \soAnc).
\\
\\
&\translation(\SoFeat \ast \kernelFuncIntra)(\translation\point, \soAnc) = (\SoFeat \ast \kernelFuncIntra)(\translation\point, \soAnc) \\ &= \sum_{\soAncj \in \SoAncGroup} \SoFeat(\translation\point, \soAncj) \kernelFuncIntra( \soAncj^{-1}\soAnc)
\\
&=( \translation \SoFeat \ast \kernelFuncIntra)(\point, \soAnc).
\end{align*}

Since both operators are \soThree{} equivariant and translation equivariant, we have:
\begin{align*}
  \rotation((\SoFeat \ast \kernelFuncInter) \ast \kernelFuncIntra)(\point, \soAnc) &= (\rotation(\SoFeat \ast \kernelFuncInter) \ast \kernelFuncIntra)(\point, \soAnc) \\
  &= ((\rotation\SoFeat \ast \kernelFuncInter) \ast \kernelFuncIntra)(\point, \soAnc),
\end{align*}
\vspace{-5mm}
\begin{align*}
  \translation((\SoFeat \ast \kernelFuncInter) \ast \kernelFuncIntra)(\point, \soAnc) &= (\translation(\SoFeat \ast \kernelFuncInter) \ast \kernelFuncIntra)(\point, \soAnc) \\
  &= ((\translation\SoFeat \ast \kernelFuncInter) \ast \kernelFuncIntra)(\point, \soAnc).
\end{align*}

Thus, the \seThree{} separable convolution is equivariant w.r.t. rotation $\rotation \in G$ and translation $\translation \in \mathbb{R}^3$, which approximates equivariance to \seThree{}.

%\section{Feature Visualization by Choice of Global Pooling Methods}
%
%Include the figures here

\section{Network Architecture and Parameters}

\begin{figure*}[h]
\vspace{-5mm}
 \begin{center}
  \includegraphics[width=0.8\linewidth]{SPNet2.png}
 \end{center}
 \caption{An illustration of the network architecture used in both ModelNet and 3DMatch experiments.}
 \vspace{-5mm}
 \label{fig:spconv}
\end{figure*}

% 32 32 64 64 128 128 256, fc: 256, stride: 2222

The network architecture used in both experiments is illustrated in Figure~\ref{fig:spconv}. Input points ($\Point \in \mathbb{R}^{\pointNum \times 3}$) are first lifted to features that are defined in the $\seThree$ space ($\SoFeat(\pointi,\soAncj): \mathbb{R}^3 \times \SoAncGroup \rightarrow \mathbb{R}$), by assigning rotation group to each point and setting its associated features to be constant 1s (denoting occupied space). 
Therefore, in the first layer, the network learns to differentiate different input points by the kernel correlation function (Equation 7 in the main text). The layer after the separable convolutional layers is an MLP layer with a symmetry function (average function) that aggregates features in the spatial dimension. We have introduced this layer as a function with implicit kernel formulation (see Equation 8 in the main text). Before the fully connected layers, a separate branch of unitary convolution takes the spatially pooled feature defined in $\soThree$, and outputs the attention confidence (see Section 3.3 in the main text). The output feature of the network can be further processed by a softmax layer in the classification task, or an $l_2$ normalization in the shape matching task. 

\section{More Implementation Details}
In the implementation of \seThree{} point convolution, we follow the design principles in~\cite{qi2017pointnet++} to compute a spatially hierarchical local structure of the points, by
 subsampling the input points with furthest point sampling and obtaining spatial local neighborhood by the ball searching algorithm. For the explicit point kernel function, we select a kernel size of 24 with kernel points evenly distributed inside a ball $\mathcal{B}_r^3$. The radius $r$ of grouping operator is set as $r^2 = d\sigma$, where $d$ is a parameter related to the density of the input points, $\sigma$ is a parameter used in the correlation function $\interCorrelation(y, \tilde{y})$ described in Section 3.1 in the main text. The kernel radius $r_k$ is set as $r_k = 0.7r$.

To achieve more effective rotation group convolution, inspired by~\cite{esteves2019equivariant}, we choose to sample the rotation group from an axis-aligned 3D model of regular icosahedral. Each face normal of the icosahedral provides the $\alpha$ and $\beta$ angles. We additionally sample three $\gamma$ angles for each face normal, each separated by 120 degrees. %The rotation is composited from the Euler angles following the xyz standard. 
The cardinality of the rotation set is thus $20 \times 3=60$. Thanks to the icosahedral symmetry, the set of rotation forms a rotation group $\SoAncGroup$ with closure, associativity, identity and invertibility. For band-limited filters, a 12-element subgroup is chosen, which transforms each element of $\SoAncGroup$ to its \soThree{} neighbors. 

It is thus worth noting that while we maintain a sparse representation for the spatial dimension of the point set, which takes online computation to find its local structure in the spatial dimension, the rotation group naturally possesses a closed grid-like structure. This greatly facilities the computation for the band-limited group convolution.

\section{Details on Experiment Setup}

\subsection{Details on ModelNet Training}

For each training object, we randomly sample 1,024 points from the input point cloud. We train our network with Adam optimizer. The learning rate is set to $0.001$ and the batch size is 16. The model is trained for 150 epochs with an exponential decay of learning rate by half for every 50 epochs. 

\subsection{Details on 3DMatch Training}

The training set of 3DMatch consists of RGB-D images in sequences from 62 indoor scenes. We denote a fused sequence of RGB-D images and its converted point cloud as a fragment of a scene. To generate training examples, we follow \cite{li2020end} to first fuse the RGB-D images into fragments. Then we convert the fragments to point clouds and select pairs that have more than 30\% overlapping region, given the ground-truth camera transformations. Therefore, each pair of the fragments comes from the same indoor scene. The point cloud patches used as input to the network are generated by gathering N=1,024 points within a support region whose radius is set to 0.4m. 

We train a Siamese network that extracts features from the source and target point clouds in parallel. Inspired by~\cite{gojcic2019perfect}, the network learns from a Batch Hard (BH) triplet loss, where negative examples are target patches ($\batch^p_i$) in a minibatch that do not correspond to the source patch ($\batch^a_i$):

\begin{align*} 
	\loss_{BH}(\batch) = \frac{1}{\batch}\sum^{\lvert\batch\rvert}_{i=1}&\text{max}(0, \lvert\lvert \sphFeat(\batch^a_i) -  \sphFeat(\batch^p_i) \rvert \rvert_2 -	 \\ 
	& \underset{{\substack{\text{j}=1\ldots\lvert\batch\rvert\\ \text{j}\neq\text{i}}}}{\text{min}}  \lvert\lvert \sphFeat(\batch^a_i) -  \sphFeat(\batch^p_j) \rvert \rvert_2 + m ),
\end{align*}

\noindent where $m$ is the margin for the triplet loss. We use a batch size of 16 for a mini-batch, which contains pairs of point cloud patches from the same pair of partially overlapped fragments. The model is trained for 30 epochs with an exponential decay of learning rate by half for every 6 epochs.  The choice of optimizer and all other hyperparameters remain consistent with the classification network.

\subsection{Inference speed.} 

We compare our model used in the experiment to the baseline models that employ similar equivariant structures regarding the inference time. Specifically, we evaluated our 20-anchor model to align with the settings in~\cite{esteves2019equivariant,kanezaki2018rotationnet}. Among the selected baselines, \cite{esteves2019equivariant,kanezaki2018rotationnet} are multi-view image networks that are SO(3) equivariant; TFN~\cite{thomas2018tensor} is an example of ``non-separable'' SE(3) equivariant network. Our network is found to be faster than all of the baselines selected, and it is significantly faster than the SE(3) equivariant framework that is not separable.

\begin{table}[h]
\small
\centering
\begin{tabular}{|c|c|c|c|c|}
\hline
Method   &  OURS-20 &  EMVN-20 & RotationNet & TFN \\ \hline
Time &   \textbf{35.4ms}   &      35.9ms   &   108.0ms  &  302.9ms            \\ \hline
\end{tabular}
\end{table}

{\small
	\bibliographystyle{ieee_fullname}
	\bibliography{paper}
}